# Towards transparent and data-driven fault detection in manufacturing: A case study on univariate, discrete time series

**Bernd Hofmann[a,*], Patrick Bründl[a], Huong Giang Nguyen[a], Jörg Franke[a]**

[a] Institute for Factory Automation and Production Systems (FAPS)

Friedrich-Alexander-Universität Erlangen-Nürnberg, Germany

*Corresponding author

*E-mail address: bernd.hofmann@faps.fau.de

*ORCID: https://orcid.org/0009-0009-0666-6149

**Abstract** Ensuring consistent product quality in modern manufacturing is crucial, particularly in safety-critical applications. Conventional quality control approaches, reliant on manually defined thresholds and features, lack adaptability to the complexity and variability inherent in production data and necessitate extensive domain expertise. Conversely, data-driven methods, such as machine learning, demonstrate high detection performance but typically function as black-box models, thereby limiting their acceptance in industrial environments where interpretability is paramount. This paper introduces a methodology for industrial fault detection, which is both data-driven and transparent. The approach integrates a supervised machine learning model for multi-class fault classification, Shapley Additive Explanations for post-hoc interpretability, and a domain-specific visualisation technique that maps model explanations to operator-interpretable features. Furthermore, the study proposes an evaluation methodology that assesses model explanations through quantitative perturbation analysis and evaluates visualisations by qualitative expert assessment. The approach was applied to the crimping process, a safety-critical joining technique, using a dataset of univariate, discrete time series. The system achieves a fault detection accuracy of 95.9 %, and both quantitative selectivity analysis and qualitative expert evaluations confirmed the relevance and interpretability of the generated explanations. This human-centric approach is designed to enhance trust and interpretability in data-driven fault detection, thereby contributing to applied system design in industrial quality control.

**Keywords** XAI, quality control, machine learning, explainable AI, human-centric, human machine interface

## 1. Introduction

Manufacturing processes generate a wide range of data across various modalities and complex relations. These data sources are essential for consistently controlling and monitoring the process, whether in-line, between individual process steps, or end-of-line, on the finished product. Especially in safety-critical components such as load-bearing or signal-transmitting joints, the comparison of actual process values against target specifications is crucial to ensure product quality. Crimping, a solderless joining method, is utilised to create a permanent, electrically conductive and mechanically stable connection between a conductor and termination [1]. This process is widespread in manufacturing, from control cabinets to automotive wiring harnesses, and is subject to strict quality requirements [2,3]. Critical quality characteristics for a reliable crimp connection are, for instance, the correct crimp height and pull out force. However, the measurement of these characteristics requires destructive testing, therefore, they can only be verified through random sampling during the production process.

To detect crimp faults such as missing strands or crimped insulation, as schematically illustrated in Fig. 1, for every connection made, the applied force during the working stroke of the crimping tool is recorded and compared to a pre-recorded reference curve [4]. In commercial crimp force monitoring systems (CFM) deviation thresholds need to be set for the in-line quality control. If the applied force exceeds these thresholds, the connection is classified as defective. Consequently, these systems rely on human-defined tolerance limits, which require extensive testing to determine acceptable product boundaries or necessitate substantial domain expertise. Additionally, the manual selection of features and boundaries may not always capture the complex patterns or unexpected defects. High variability in manufacturing data, resulting from material fluctuations, environmental influences, and human factors, further complicates the initialization and interpretation of these statistical-based monitoring methods.

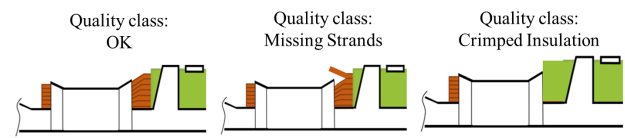

**Fig. 1.** Schematic quality classes of crimp connections.

Machine learning (ML) presents a promising solution [5], as it can detect high-dimensional patterns in production data that remain hidden to human observation. Unlike systems that base on human-defined feature and threshold selection, ML algorithms are capable of autonomously identifying highly complex boundary rules allowing for more adaptive and fine-grained fault detection. Despite their demonstrated potential in the domain of crimping [3,6–8] these approaches often lack transparency in their decision-making process. This not only impairs acceptance by domain users, but mainly hinders their widespread adoption in serial applications [9].

This challenge presents a fundamental trade-off in industrial quality control. On the one hand, human-defined feature extraction and threshold-based monitoring are transparent and interpretable but require extensive manual tuning and may miss complex defect patterns. On the other hand, data-driven approaches improve detection accuracy but have an explainability gap, making it difficult to adapt in safety-critical production environments.
To address this issue, this paper proposes a methodology for industrial fault detection, which is both data-driven and interpretable. The approach is evaluated in the domain of crimping, but designed to be adaptable to a broader range of manufacturing processes, making it applicable across various industrial settings.

## 2. Related work

Given the importance of transmitting power or signals under strict safety requirements, implementing a reliable and transparent quality control system is essential. This section examines how CFM is applied in automated series production, explores advances in data-driven approaches, and discusses the use of explainable AI (XAI) in time-series analysis to enhance the transparency and interpretability of data-driven models.

*2.1 Crimp force monitoring and limitations*

To analyse the crimp force curve and detect defects in the connection, distinct zones have been defined, as different defects impact different phases of the force curve [4,10–12]. Given the wide range of interpretation approaches, the definition illustrated in Fig. 2 is established for this study.

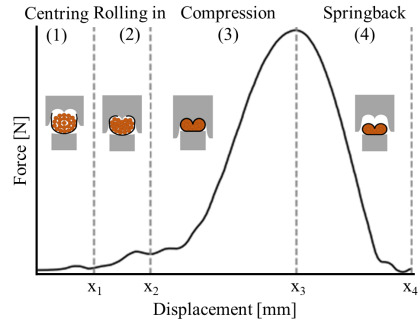

**Fig. 2.** Zone interpretation of a crimp force curve.

The crimp force curve can be segmented into four phases - Centring, Rolling in, Compression, and Springback. The first phase Centring (1) begins when the terminal flanks first touch the crimp stamp. The crimp flanks are aligned along the entry bevels of the stamp. After (1), the crimp flanks begin to be rolled in by the crimping tool. This is the start of the Rolling in (2) phase. The crimp flanks are guided by the tool, bent around the stranded wire and plastically deformed. Due to the bending resistance of the terminal flanks, a slight and brief increase in force may be recognized. The Compression (3) section is the segment with the highest recorded increase in force and begins when the ends of the crimp flanks are bent around the strand and touch the individual wires. A significant increase in the force curve occurs, when the wires and the surrounding flanks are compressed and plastically deformed. When the stamp reaches its lowest point, the final phase of the crimping operation, the Springback (4), begins. The elastic deformation of the crimped connection results in a springback and the tool begins to detach from the crimp connection. This can be recognized in the process curve by a rapid drop in force.

By analysing the applied force in the curve zones, CFM systems can detect process deviations [13]. Fig. 3 illustrates how a explementary CFM system [14] compares the recorded force curves against predefined reference boundaries. The logic behind such systems relies on manually defined tolerance boundaries set by domain experts, with fixed minimum and maximum limits. The recorded force curve (black) is compared to a pre-recorded reference curve (red), which is initially validated through destructive testing methods. If all areas under the curve (purple, blue and yellow surface) fall within the specified boundaries, the crimp connection is classified as acceptable (OK). If any area exceeds these limits, the connection counts as defective (NOK). In other CFM systems, the specified areas or analysed features may vary depending on the manufacturer and the applied methodology. For instance, the four zones delineated in Fig. 2 can be applied, or the peak force or slope of the curve can be utilized as a quality feature. One of the primary advantages of such a rule-based method is its inherent interpretability, whereby operators can directly recognise which section of the force curve has exceeded the predefined limits, thus allowing them to deduce the cause of a detected error with expert knowledge. However, a significant limitation of this approach is that it relies on a small set of manually defined features, such as the three areas under the curve, rather than utilising the full range of available data points. This restricted feature selection and threshold setting can lead to unintended consequences, including the triggering of false alarms due to minor process variations, and the inability to detect actual defects that cause only subtle deviation in these features. This problem arises because no generalisable patterns are learned that effectively distinguish between an OK and NOK connection as the selection of features and threshold values is done manually rather than through data-driven optimization. Instead, the reference curve is limited in its capacity to represent the variable nature of the domain of crimping, due to the fact that it is representative only to a small set of instances. In summary, the state-of-the-art in CFM facilitates real-time defect detection through manual and interpretable feature definitions and threshold-based decision rules. However, this approach necessitates frequent recalibration and exhibits limited adaptability, leading to time consuming and costly setup procedures. These constraints highlight the need for data-driven approaches that can automatically learn complex discriminatory patterns from crimp force curves.

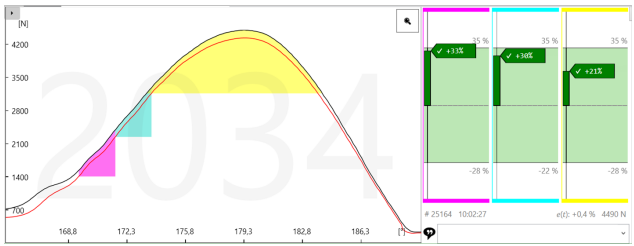

**Fig. 3.** The quality characteristics "areas under the curve" (purple, blue and yellow surface) from the recorded force curve (black) are compared to the surfaces of a prerecorded reference curve (red). An OK connection must not exceed the tolerance boundaries (right side).

*2.2 Data-driven crimp force monitoring*

Accurately describing the relationship between the features of the crimp force curve and the joint quality is challenging due to its complex behaviour. Therefore, researchers have investigated data-driven techniques to improve fault detection during the crimping process. ML has the ability to uncover subtle patterns in force curves that manually selected features might miss. Several studies have applied both supervised and unsupervised learning methods to classify crimp quality based on the force curve.

Meiners et al. [3,6] demonstrated the effectiveness of deep learning techniques for process curve monitoring in manufacturing. They employed one-dimensional convolutional neural networks (1D-CNNs) for mechanical joining processes like screwing, pressing, and crimping. Their study achieved high accuracies ranging from 86.9 % to 100 % in multi-classification tasks across these processes. For anomaly detection, they used autoencoders and variational autoencoders, achieving recall values up to 99.02 % for critical anomalies in crimping.

Song et al [7] introduced regional selective data scaling to generate synthetic abnormal data, addressing the challenge of limited access to labelled data in quality control for wiring harness crimping manufacturing. Their approach demonstrated a compelling alternative to traditional anomaly detection methods and their benchmarked CFM system. The proposed system achieved an average accuracy of 99.95 %.

Nguyen et al. [15] applied a combination of data-driven CFM and optical inspection to enhance quality control in the crimping process. They integrated ML techniques, using one-dimensional convolutional autoencoders for anomaly detection and 1D-CNN for multi-classification tasks. Their optical inspection system used a multi-camera setup to capture 360-degree views of stripped and crimped wires, achieving up to 99.7 % accuracy in both fault classification and anomaly detection.

Hofmann et al. [8] emphasized the potential of ML for improving CFM in wiring harness production with focus on the use of diverse data. They pointed out the limitations of conventional monitoring methods and the need for adaptable, data-driven solutions. By utilizing datasets from various machines and wire cross-sections, they applied supervised learning techniques and achieved accuracies up to 95 %, surpassing today's systems. Their research demonstrated the robustness and effectiveness of ML in managing diverse manufacturing conditions.

These prior studies have shown the effectiveness of ML for fault detection in crimping. However, most of the models used are inherently complex and fall under the category of black-box algorithms, which are characterized by their lack of interpretability in their decision-making processes [16]. This missing transparency is particularly problematic in safety-critical scenarios, such as joints in automotive or aerospace applications, where interpretability is mandatory for validation [17], operator trust [18], and subsequent actions like root cause analysis [19]. State-of-the-art CFM systems visually display deviation from a reference curve, which operators can directly interpret. In contrast, current ML-based solutions do not inherently provide such transparent feedback. In the use case of crimping, it is essential to understand which parts of the force curve led to the classification of a particular quality class to quickly implement a counter measure. This fundamental aspect is a core component of the principle Jidoka, a main pillar of the Toyota Production System [20]. To analyse manufacturing data effectively and initiating root cause analysis, it is essential for operators to not only recognize that a process anomaly occurs but also

"[…] why the AI/ML model made a certain decision" [21]. Moreover, when a data-driven system makes a mistake (a false alarm or a missed defect), it is challenging to diagnose the cause if the internal logic is obscured. These concerns necessitate the integration of XAI techniques. XAI methods aim to address this gap by providing transparency and interpretability in the decision-making process of a ML model [22]. Therefore, an effectively integrated XAI methodology in a fault detection system could offer a user-centric pathway to not only identify process anomalies but also understand the decision-making process behind the model to quickly implement countermeasures.

*2.3 Explainable AI for time series data*

The crimp force curve can be described as a univariate, discrete time series where the sequence of values (force or signal) varies over time or distance (rotation angle of the press or distance travelled by the crimp stamp). While XAI has seen extensive development in domains like computer vision, time-series data poses unique challenges and there has been increasing research on adapting XAI for time-series classification and anomaly detection [23]. Various taxonomies have been developed to categorise explainability techniques into meaningful groups [16,21]. One common categorisation is based on the scope of explanations, with a distinction made between holistic global explanations, which provide an overarching view of a model's decision-making process, and local explanations, which focus on individual instances and the corresponding model results. A further classification considers the phase in which interpretability is applied, with ante-hoc interpretations referring to methods used to analyse black-box models before training, and post-hoc interpretations referring to techniques used to analyse models after training. According to Theissler et al. [23] and in the context of time series data, explanatory methods can additionally be categorised into the type of explanation they return. Time-based explanations focus on specific points within a time series, while subsequence-based explanations refer to sub-segments of the series. The third option are instance-based explanations adopting the entire time series as the basis for the explanation. Additionally, the evaluation of the explanation can be broadly categorised into quantitative approaches, which focus on evaluation metrics and model performance, and qualitative approaches, which rely on subjective human judgements [23].

These classifications facilitate precise differentiation between available approaches, thereby providing an overview from which a selection can be made for a specific use case. Nevertheless, a key conclusion of the comprehensive study by Theissler et al. [23] is that, while a general modelling and explanation approach is desirable, domain-specific solutions are required. The effectiveness of an explanation depends on the user's perception and the specific characteristics of the domain, which emphasises the need for use case specific interpretation methods.

Consequently, interpreting a fault detection system based on a time series data with a data-driven approach presents a significant challenge. Single data points of a raw sequence, such as the crimp force curve, lack intuitive features that are interpretable by humans. This deficit makes it difficult to understand the explanations provided. Moreover, the manner in which the explanatory information is communicated to factory operators plays a significant role in ensuring that the explanations are both valid and interpretable. Basic plotting methods like bar charts can fulfil visualisation requirements, as long as attributes carry interpretable meaning. However, simply highlighting the most relevant features is not a sufficient approach for complex data patterns like in time series. As illustrated in the bar chart plot from the SHAP library [24] in Fig. 5, a limitation lies in its lack of intuitiveness and disconnection from the actual sequence of input features. Similar to a single pixel in an image, an individual data point in a time series may not be informative. In Fig. 4, the importance values for the nine most relevant features range from 0.01 to 0.02, while the remaining 491 data points collectively account for 0.66. Consequently, there is no single critical data point. Instead, significant regions may exist within the time series (subsequence-based explanations). For instance, important areas may be found between the datapoints of $\in [110,120]$ or $d \in [250,260]$. Furthermore, the absolute importance values are not inherently meaningful to factory operators, and therefore cannot be interpreted directly.

In summary, the continuous nature of the time series, in conjunction with their domain-specific characteristics and the absence of a general modelling and explanation approach, complicates the development of an interpretable data-driven fault detection system. These challenges emphasize the need for further research to enhance the explainability and usability of data-driven quality control systems in industrial setting and practical applications.

**3. Dataset**

Due to the lack of publicly available, domain-specific data, the dataset used in this work was independently collected, pre-processed, and labelled. The dataset consists of two different conductor-terminal pairings. The first copper conductor has 16 strands and a wire cross-section of 0.50 mm², the second 12 strands and a wire cross section of 0.35 mm². For this study a total of 1617 crimp force curves $F(d)$ were recorded, of which 845 belong to the cross section of 0.5 mm² and 772 to 0.35 mm². The force curve can be described as:

$$F(d) = \{F0, F1, F2, \dots, Fn\} \quad (1)$$

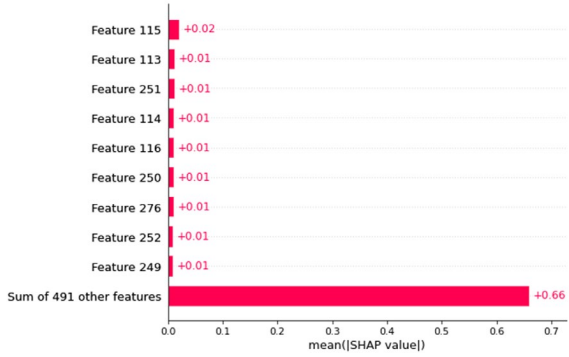

**Fig. 4.** Explementary bar plot visualisation from the SHAP-library.

The data utilised in this study is openly available [25]. The data aggregation and pre-processing steps are described in the following sections.

*3.1 Data aggregation*

The force curves were recorded on a semi-automatic crimping machine [26], which automatically feeds the terminals and executes the crimping process, but requires manually insertion of the conductor. Since the machine is neither cutting nor stripping the conductor, these steps were performed with a manual stripping tool before. During the crimping process a force sensor [27], which is directly placed on the connection rod of the press, continuously records the applied force. For every crimp connection made one crimp force curve is recorded. Along with the acceptable OK quality class, two common defective classes, Missing Strands and Crimped Insulation, were intentionally produced to enrich data variability. Table 1 summarizes the dataset and explains how the quality classes were prepared. The recorded force curves are saved in CSV files with a unique ID and manually labelled. The Missing Strands class represents two common issues in series production of automated crimping machines: incorrect blade settings that cut too deep, resulting in cut strands, and conductors deformed by the weight of cable reels over time, leading to non-round conductors. The Crimped Insulation class simulates issues such as incorrect blade settings or improper feeding of the conductor, either not far enough into the blade station or too far into the crimping station, indicating incorrect stop settings.

*3.2 Data pre-processing and preparation*

Pre-processing of data is a crucial for ML algorithms, since their learning ability is directly affected by the quality of the input data [28]. Preparation steps like cleaning and scaling as well as the transformation of the raw input features makes the data more efficient in processing [29]. Scaling the data is especially important for ML algorithms that use gradient descent as an optimizer, such as neural networks, since it can help gradient descent converge more quickly towards the minimum [28].

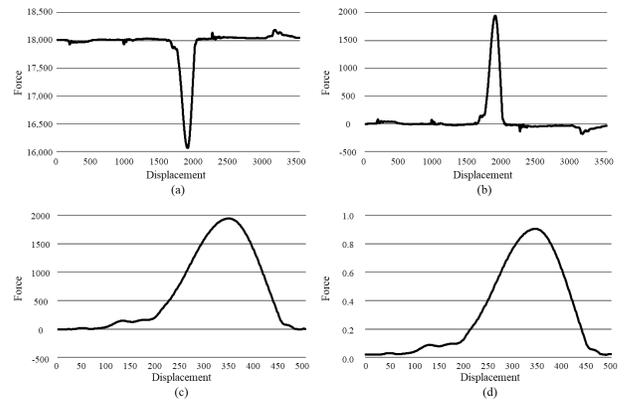

**Fig. 5.** Data preparation steps from the raw recorded force (a) over ensuring data comparability (b) and windowing to the region of interest (c) to the scaled curve (d).

Russel et al. [30] categorized the handling of input features into two approaches for predicting the quality in resistance spot welding: feature-based and raw sensing-based. In the feature-based approach, features are manually defined using domain knowledge, whereas the raw-sensing approach utilizes the raw time series data directly. In crimping, a manually defined feature could be the peak force value, which indicates whether the crimp connection is good or bad due to missing strands. Another example are the areas under the curve, as presented in section 2.1, which are commonly used in state-of-the-art CFM systems. However, this approach can result in significant information loss, particularly in highly complex processes such as crimping. Additionally, the different wire cross-sections might cause overlapping across the various quality classes of these selected features. A further disadvantage of this method is that it offers a less transparent decision making process than the raw sensing approach. This is due to the fact that the abstraction of the raw data into potentially complex mathematical features may lack physical explanations, making them less informative for understanding the dynamics of the crimping process. To maintain maximal information from the time series, this study adopts the raw-sensing approach, which feeds the entire normalized force curve into the model. The ML algorithm can learn and extract the relevant features by its own to predict the quality classes from a data driven perspective. Although the pre-processing does not involve feature engineering, the data is still prepared in several steps to ensure data quality for the model. The preparation steps are illustrated in Fig. 5. After recording the raw sensor data, the curves are inverted, as the force sensor is mounted upside down. Afterwards, the curve is set to zero to ensure comparability with curves from potential other machines as the absolute sensor values are defined manually. Subsequently, the 3567 data points from the raw curve are reduced to the region of interest, defined as the window, where the crimp stamp first contacts the terminal and ends when it no longer touches it. The sensor records data points before and after this window, but this data is irrelevant to assessing crimp quality, resulting in a total of 500 data points. The crimp force curve shortened to the region of interest $F_i(d)$ can be defined as:

$$F_i(d) = \{F_i0, F_i1, F_i2, ..., F_i500\} \qquad (2)$$

**Table 1.** Overview of the dataset and quality classes.

| Quality class | Image of stripped conductor | Explanation of the preparation procedure | Quantity |
| --- | --- | --- | --- |
| OK | 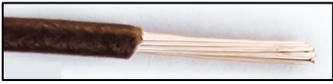 | The conductor was stripped with OK defined length. The manual cutting and automatic crimping process was conducted correct as well. | 805 |
| Missing Strands | 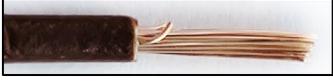 | The conductor was stripped with OK length. One to three strands were cut off manually after stripping. The crimping process was conducted correct. | 515 |
| One |  | One strand was cut off manually. | 172 |
| Two |  | Two strands were cut off manually. | 292 |
| Three |  | Three strands were cut off manually. | 51 |
| Crimped Insulation | 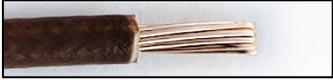 | The conductor was stripped with a defined NOK length. The manual cutting and automatic crimping process were conducted correct. | 297 |

The next step involves min-max scaling of the curve using the scikit-learn library [31], with the force curves linearly scaled into a fixed range, where the largest data point is designated as the maximum value and the smallest as the minimum value [32]. Finally, the dataset is divided into training and testing set, with 80 % (1,293 curves) of the data used for model training and 20 % (324 curves) reserved for testing.

## 4. Transparent and data-driven fault detection

The primary objective of this study is to design a transparent data-driven fault detection system, which comprises three key components: a fault detection algorithm, an explanation logic of the decision-making process, and a visualisation technique suitable for factory operators. The Random Forest (RF) algorithm [33] has been selected on the basis of its superior performance in prior studies [8], and was implemented from the scikit-learn library [31]. The second key component, explaining the decision-making process of the fault detection, needs to be able to uncover the relationship between the input feature values and the predicted quality class. The Shapley Additive Explanations (SHAP), as proposed by Lundberg et al. [24], was utilised as a model-agnostic, post-hoc explanation method. To facilitate comprehension, a subsequence-based explanation approach with overlaying pipes and a normalized colour gradient was selected to direct user attention to the most salient features. The transparent quality control system is designed to function in two distinct capacities: firstly, to detect faults, and secondly, to provide the user with a detailed explanation of the reasoning behind its conclusion. The predictions of the fault detection are evaluated by various metrics relevant in multi-class classification tasks. The explanations and visualisation are evaluated using a quantitative and qualitative approach.

### 4.1 Data flow and system inference

The data pipeline consists of several key stages, starting with data preparation as described in section 3, followed by processing through three core system components: fault detection, explanation, and visualization Fig. 6 illustrates the inference process from the recorded crimp force curve to the final visualisation of the quality control system, including the relationships among the system components. First the crimp force curve is recorded during the manufacturing process and stored in a CSV file. The data is then converted into a Numpy array and the region of interest is extracted from the curve. The pre-processed array is fed into the black-box model, which categorises the instance into one of the three quality classes. Afterwards, an explainer algorithm is applied to interpret the model's decision. The explainer computes and assigns importance scores to each datapoint of the force curve and stores the result in a Numpy array. Afterwards, the array is sliced into the four predefined phases defined in literature, and based on the aggregated importance scores a phase importance value is calculated for each curve phase. The computed values are normalized and a colour gradient is generated. In the final visualisation the pre-processed force curve is plotted and overlaid with highlighted and colour-coded pipe segments. The most critical curve phase and the specified quality class is displayed.

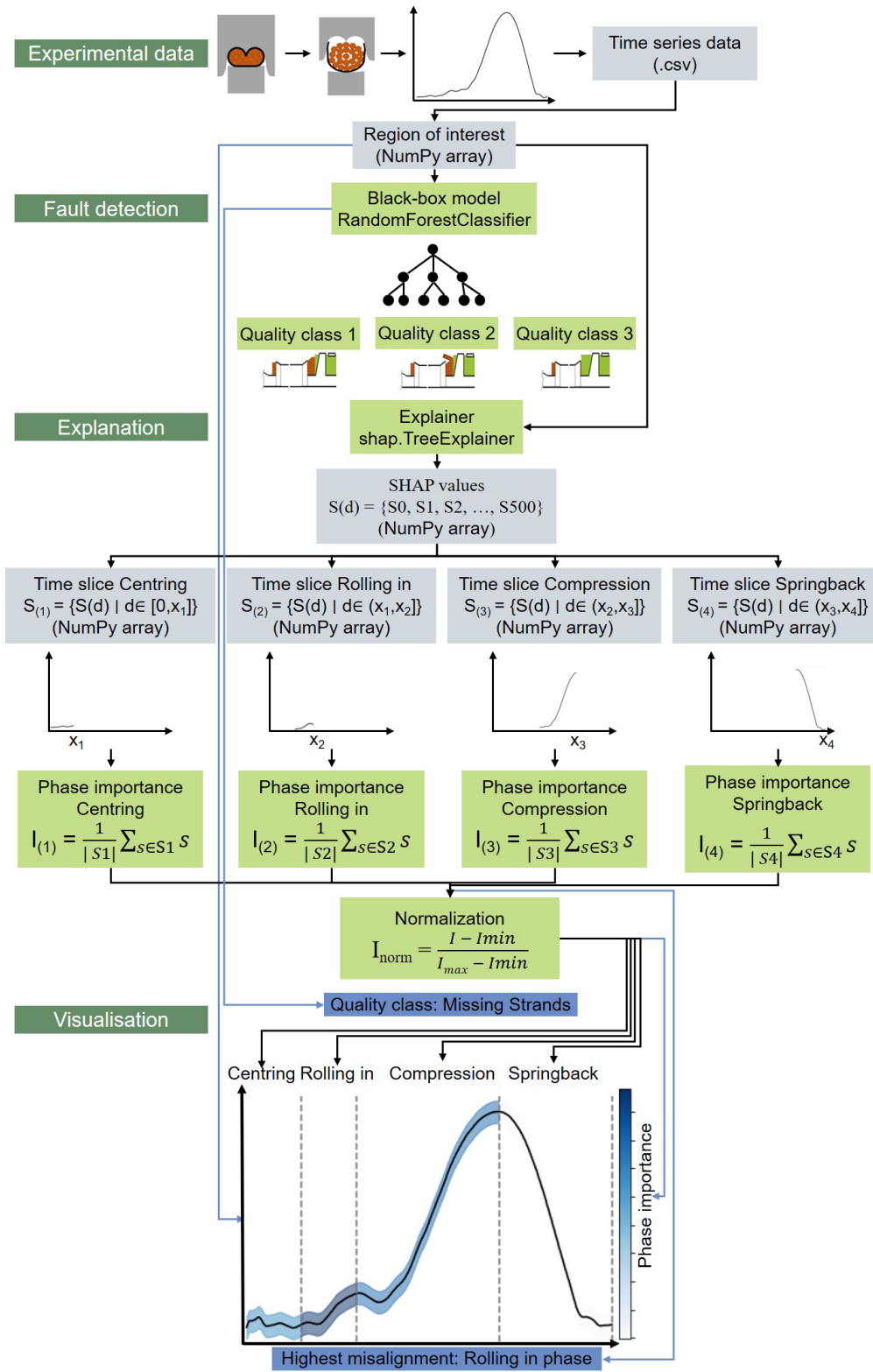

**Fig. 6.** Data flow and inference of the presented transparent data-driven fault detection system

*4.2 Random Forest for fault detection*

Prior research [3,6,8,15] has already demonstrated that supervised learning algorithms have the potential to substitute or at least supplement state-of-the-art CFM systems in the field of crimping. Thus, the primary focus of this research is not to improve the prediction accuracy of the fault detection, but to propose a transparent system. Fault detection is categorized as a multi-class classification problem, and the black-box model RF [33] is employed as a supervised learning approach to predict the quality classes of the crimp connections. The RF algorithm serves as a meta estimator that constructs multiple decision tree classifiers on various sub-samples of the dataset and utilizes averaging to boost predictive accuracy [33]. This method is known for its robustness against overfitting and has shown superior performance in previous research [8]. Given that model performance is significantly influenced by the selection of hyperparameters, Grid Search [34] from the scikit-learn library [31] was implemented to identify the optimal hyperparameters. This approach employs integrated k fold cross validation, which divides the training set into multiple folds to ensure that the model generalizes well to unseen data and evaluates the average performance across each of the k iterations [34]. The defined parameter grid is presented in Table 2.

**Table 2.** Overview of the hyperparameter grid.

| Hyperparameter | Values |
| --- | --- |
| n_estimators | [50,100,200,300,400] |
| max_depth | [None, 5,10,20,30] |
| cv | 5 |

*4.3 SHAP for explanations*

The SHAP approach, described by Lundberg et al. [24], assigns Shapely values as a unified measure of feature importance, and hence attributes each feature with an importance value for a particular prediction. This includes an analysis of the interaction of the features with each other, whereby permutations are utilised to measure the change in the prediction. To interpret the Shapley values, an analysis of both their absolute value and their direction is necessary. The absolute value indicates the degree of influence of the features, while the direction shows a positive or negative influence on the prediction for each feature. With regards to the input features of this case study, each of the 500 datapoints of the force curve is assigned an importance value that states how it contributed to each of the quality classes. The shapley values of a force curve $S(d)$ can be expressed as follows:

$$S(d) = \{S0, S1, S2, ..., S500\} \quad (3)$$

From the SHAP library [35] the TreeExplainer [36] is used for local instance explanation and global understanding of the ensemble tree model.

*4.4 Slicing of the force curve*

As described in section 2.1 a crimp force curve can be divided into four phases to describe its physical behaviour. The slicing of a time series into contiguous subseries is described by Mujkanovic et al. [37] as Time slice mapping. In regards of the explanation return subsequence-based is the proposed terminology of Theissler et al [23]. In time series data, individual data points are typically part of a continuous sequence and may not hold distinct significance when viewed in isolation. Instead, their value lies in the patterns and relationships formed across multiple points over time. For this use case the delineation of the individual crimp curve phases is defined as in section 2.1 and can be described as follows:

$Centring: (d \in [0, x_1])$, $Rolling\ in: (d \in (x_1, x_2])$,
$Compression: (d \in (x_2, x_3])$, $Springback: (d \in (x_3, x_4])$

with the phase boundaries defined as:

$$x_1 = 75, \ x_2 = 150, x_3 = 345, x_4 = 500$$

Consequently, the Shapely values of a force curve phase is described as:

$$S_i = \{S(d) | \ d \in (x_{i-1}, x_i]\} \quad (4)$$

for $i = 1, 2, 3, 4$

To express the significance of a complete curve phase the average Shapley value of the elements within the phase $I(i)$ is calculated:

$$I(i) = \frac{1}{|S_i|} \sum_{s \in S_i} s \quad (5)$$

for $i = 1, 2, 3, 4$

*4.5 Visualisation*

Another important attribute of human-centric systems is the visual representation [18]. Hence, the third key component of the system consists of the visualisation technique of the explanations. Without a proper visualisation, it can still be difficult for a domain user to understand the decision-making process of the fault detection model. In image-based contexts, attribution heatmaps are commonly used to highlight the most relevant pixels [38]. However, since time series data is primarily represented using line plots, this approach cannot be directly applied but is still indirectly transferable. In the proposed approach, the following principles are applied to ensure an interpretable visualisation:

- Feature Integration: The highlighted features must remain connected to the actual time series sequence. The Shapley values should be integrated into the line plot of the force curve rather than visualized separately.
- Subsequence-based explanation: Instead of displaying importance values for individual data points, the visualization aggregates the Shapley values over curve phases, aligning with the explanation approach of the physical behaviour in literature.
- Colour Coding: Since absolute Shapley values are not directly interpretable for operators, a color-coded representation is applied. The importance values of the four curve segments are normalized to generate a colour gradient, which is then used to shade the surrounding pipe in the force curve plot. This method effectively guides the user's attention toward significant curve segments rather than isolated data points.
- Quality Classification: The quality class predicted by the fault detection model is displayed above the plotted force curve. Compared to binary anomaly detection systems, the supervised learning algorithm used in this approach provides superior information for subsequent steps, such as troubleshooting.

*4.5 Evaluation method*

The system is composed of three subcomponents, each requiring specific evaluation methods. The evaluation approach of this study is illustrated in Fig. 7. The assessment of the fault detection system's capability to correctly classify quality classes is conducted using a confusion matrix, along with derived performance metrics such as accuracy, precision, recall, and F1-score.

To quantitatively evaluate the identified explanations, the first step involves analysing the computed importance values of the base model to determine which curve phases were most influential in classifying the two faulty quality classes. To assess whether the curve phases identified were indeed relevant to the fault detection decision, the selectivity metric proposed by Solís-Martín et al. [39] is applied. Selectivity measures the impact on model prediction accuracy when individual features are removed. If the features identified by the base model analysis are truly relevant, their removal should lead to a decline in accuracy. Conversely, if the features are redundant or non-essential, the accuracy should remain largely unaffected. For the evaluation, a perturbation analysis is performed, in which specific curve phases are either replaced with a fixed replacement value, such as zero or the mean value [40], or be removed entirely [41]. Furthermore, Hooker et al. [41] propose that the prediction model should be retrained for perturbation analysis. Without retraining, the model cannot learn that the replacement value α is uninformative. The prediction of the base model might conclude an anomaly otherwise. For this study three replacement values α are defined: zero $α_0$, random $α_R$ and remove $α_X$. Consequently, the data base of the selectivity analysis consists of 42 datasets and individual models (3 replacement values $α$ and 14 combinations of modified areas). While each of the 42 models is trained on a distinct manipulated dataset, the training conditions of the base model are maintained to enhance comparability. Finally, for the evaluation, the performance of the 42 manipulated models will be benchmarked against the base model, which was trained without manipulated data, to facilitate a quantitative performance analysis.

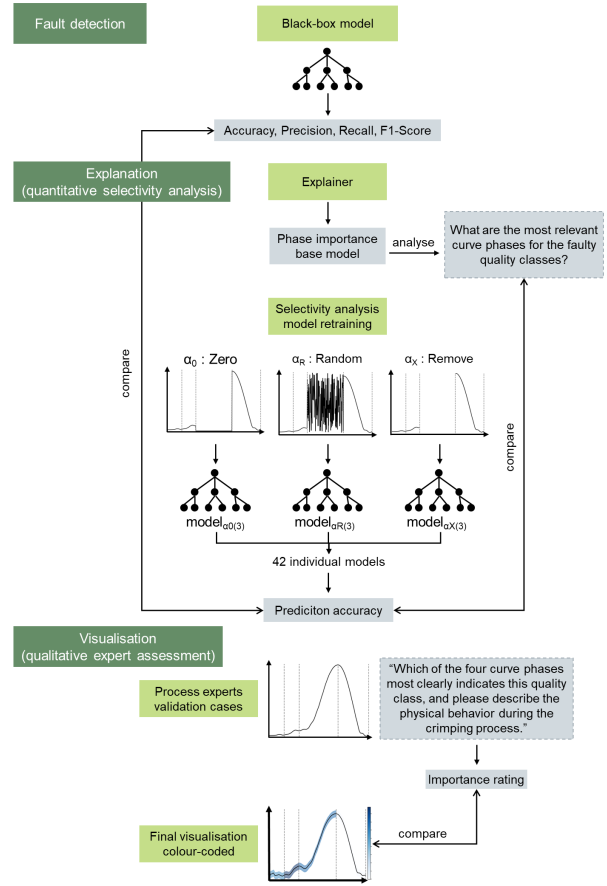

**Fig. 7.** The quantitative selectivity analysis evaluates the explanation and the qualitative expert assessment evaluates the visualisation.

The visualisation is evaluated separately, as the importance values are translated and normalized to ensure interpretability. The raw values lack an intuitive meaning for the operators, and therefore, there can be a gap between the importance value and the colour-graded visualisation. The evaluation is conducted using two validation cases. Process experts are independently asked to identify the most relevant phases for a specific quality class. Specifically, they are presented with a crimp force curve labelled with a specific quality class and asked: "Which of the four curve phases most clearly indicates this quality class, and please describe the physical behaviour during the crimping process." For each shown quality class the experts rate each segment on a scale from 0 to 3, where 0 indicates the segment has no impact or influence, and 3 expresses a significant impact. These expert ratings are then compared with the visualisation results and qualitatively analysed. This step is essential, as unlike the quantitative evaluation approach, this method directly assesses the final visualization output presented to the domain user.

## 5. Results

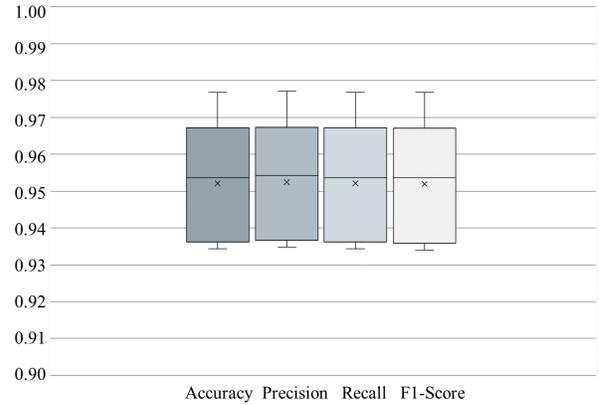

**Fig. 8** Accuracy, Precision, Recall and F1-Score of the 5 folds of the training phase

All results were obtained on the same operating system, and metrics were computed under identical conditions. To minimize human bias, random data splitting was applied, and the metrics were calculated using the scikit-learn library [31].

### 5.1 Fault detection

The training of the fault detection model was conducted using 5-fold cross-validation. During the training phase, the hyperparameters were optimised using Grid Search [34], resulting in max_depth of 20 and n_estimators of 150. Consequently, the classifier comprises 150 decision trees, each with a maximum of 20 splits from the root node to the leaf node. The 5-fold cross-validation yielded accuracy scores of 0.934, 0.953, 0.957, 0.937, and 0.976 during training phase as depicted in Fig. 8. The performance of the fault detection algorithm is evaluated by the confusion matix of the test set, which is illustrated in Fig. 9.

The matrix reveals that a total of 6 out of 161 instances classified as OK were misclassified as defects from the base model. In manufacturing context, false negatives result in financial costs due to a higher scrap rate. Conversely, only 4 out of 163 defect instances were misclassified as OK (false positives). Although this could pose safety concerns, it is noteworthy that in all four cases, the preparation step to provoke this quality class was only slightly manipulated towards a NOK state that even the current CFM system was not able to detect either. Additionally, 3 out of 163 defect instances were misclassified within the defect classes, which would result in sending incorrect information to the operator. Overall, only 4 % of all instances were misclassified by the base model, demonstrating its applicability for the fault detection task. These findings align with the high potential demonstrated in previous studies for classifying quality classes of crimp connections, as stated in section 2.2. Overall the model achieves an accuracy of 0.959, and macro average precision of 0.958, recall of 0.952 and F1-score of 0.955.

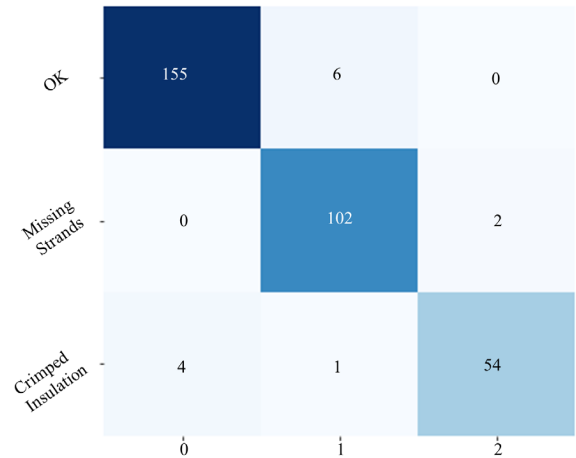

**Fig. 9.** Confusion matrix of the fault detection system.

### 5.2. Quantitative evaluation of the explanations

In addition to providing a precise classification of individual quality classes, a fault detection system also needs to be transparent in its decision-making process. Consequently, the system must highlight the key features that influenced the model's classification decisions with an interpretable approach. To analyse how the base model distinguishes classes, Shapley values are computed. As described in section 4.3, each instance consists of an array of Shapley values $S(d)$ with 500 data points. The array is sliced into the four curve phases $S_{(1)}, S_{(2)}, S_{(3)}$ and $S_{(4)}$, and for each segment, the average importance value $I_{(1)}, I_{(2)}, I_{(3)}$ and $I_{(4)}$ is calculated. Therefore, for each of the four curve phases a total of 324 importance values are calculated for the base model. Table 3 summarizes the Shapley values, which represent the average importance value of the datapoints within the four curve phases for each instance of the test set. The values are categorized per quality class, and the highest (green) and lowest (red) values per class are highlighted. For the quality class Missing Strands it stands out that the most relevant phase of the force curve is (2). Area (3) also contributes, but (1) only to a smaller extent. Phase (4) appears mainly irrelevant. However, due to the high variability it can be interpreted that not all instances follow this overall conclusion. Even though the main contributor is phase (2), it can be derived that a significant proportion of instances were classified into this quality class without area (2) being the primary factor. Some instances may rely more heavily on area (3) or to a smaller extent on area (1). Interpreting the sub-quality classes One Missing Strand, Two Missing Strands and Three Missing Strands, it is evident that the relevance of area (3) increases for the classifier the more strands are missing.

**Table 3.** Importance values of the base model per phase and quality class.

| Curve phase | Missing Strands (total) | One Missing Strand | Two Missing Strands | Three Missing Strands | Crimped Insulation |
|---|---|---|---|---|---|
| Centring (1) | 0.000828 ±0.000791 | 0.001040 ±0.000649 | 0.001366 ±0.000728 | -0.000178 ±0.000443 | **0.000400** ±0.000529 |
| Rolling in (2) | **0.001959** ±0.001152 | **0.001934** ±0.000969 | **0.003662** ±0.001183 | 0.001028 ±0.001148 | 0.000903 ±0.000535 |
| Compression (3) | 0.001677 ±0.000823 | 0.001176 ±0.000617 | 0.002940 ±0.000660 | **0.003038** ±0.000604 | **0.002709** ±0.000793 |
| Springback (4) | **0.000269** ±0.000340 | **0.000467** ±0.000360 | **0.000348** ±0.000283 | **-0.000058** ±0.000073 | 0.000420 ±0.000280 |

Again, the importance of phase (1) also contributes, but consistent with the findings before, area (4) appears to have minimal relevance over all three sub-classes. The distribution of the Crimped Insulation quality class indicates that segment (3) has clearly the greatest influence on this error pattern. Phase (1) and (4) seem to play no significant role in the model's prediction decision of this quality class. In contrast to the Missing Strands class area (2) has only a supportive impact.

Overall it can be derived, that phase (4) of the force curve is not a relevant feature for the fault detection. Phase (3), however, appears to be very informative for both quality classes. For the quality class Missing Strands with only one or two strands missing Area (2) contributes strongly, while area (1) still supports.

The next step involves verifying the results of the phase importance analysis and assessing whether the phases with the highest importance values indeed have a significant impact on the decision-making process of the fault detection. If the explanations provided by the model can be validated, an operator may gain confidence in the fault detection system, trusting that it not only classifies the correct quality class, but also accurately highlights the relevant features which caused this classification. For that, a perturbation analysis is conducted. The performance of the models trained on the manipulated datasets will be compared to the performance of the base model. The accuracy score of 0.959 will serve as a benchmark. From the conclusions from phase importance analysis it is expected that a fault detection model trained with modified curve phase (3) will perform worse compared to the base model. Whereas a manipulated phase (4) should have no significant impact. The models are all trained with the same methodology as the base model to improve comparability.

Table 4, Table 5 and Table 6 summarise the results of the perturbation analysis. The tables show the test accuracy scores of the manipulated models with the three replacement values. The lowest performance score, indicating that this manipulated phase has the highest impact, is highlighted in green, whereas the highest performance score, signifying the lowest impact, is highlighted in red.

It can be noted that the accuracy of the base model (0.959) cannot be maintained when a single curve segment is manipulated. However, which of the area is manipulated plays a significant role. For example, the accuracy remains relatively close to the base model when phase (1), (2) or (4) is modified. In contrast, the accuracy only decreases sharply for area (3), where the accuracy drops significantly to 0.861 across all replacement values. This observation aligns with the expectations outlined in the phase importance analysis, which indicated that modifying phase (3), an important region for distinguishing both fault classes, would lead to a substantial accuracy reduction. Conversely, the minimal impact observed when manipulating area (4) is consistent with its low importance for classification performance. Additionally, it can be noted that the replacement values do not appear to significantly impact the overall test results.

When two phases are modified, several findings are consistent with earlier observations, but additional insights can be derived. As expected, the accuracy of the base model cannot be achieved with additional manipulated data points. Similarly, the replacement values continue to have only minimal impact on the results, with some exceptions. For instance, when phase (1, 3) is manipulated, the accuracy score for the highest replacement value ($\alpha_0$ and $\alpha_R$) is approximately 2 % lower than for $\alpha_X$. The highest overall accuracy scores are observed when phase (1, 4) and (2, 4) are manipulated. These results suggest that modifying these combinations of phases have no significant impact, and therefore, are not highly informative for the fault detection. In contrast, the lowest accuracy scores are found which involve manipulated combinations that include phase (3). This again aligns with the expectation that phase (3) is very relevant for the fault detection. Interestingly, the lowest score occurs where phase (3, 4) is modified. While phase (4) was previously identified as unimportant, its combination with (3) leads to a notable accuracy drop. This could be attributed to the complete absence of peak value and slope information, which is critical for detecting features such as missing strands. Additionally, phase (3) and (4) include more data points than the other segments, which may influence the outcome and should be acknowledged as a potential limitation of this study.

**Table 4.** Accuracy of the selectivity analysis with one phase manipulated.

| $\alpha$ | (1) | (2) | (3) | (4) |
|---|---|---|---|---|
| $\alpha_0$ | 0.948 | 0.954 | 0.861 | 0.954 |
| $\alpha_R$ | 0.948 | 0.954 | 0.861 | 0.954 |
| $\alpha_X$ | 0.954 | 0.954 | 0.861 | 0.954 |

**Table 5.** Accuracy of the selectivity analysis with two phases manipulated.

| $\alpha$ | (1,2) | (1,3) | (1,4) | (2,3) | (2,4) | (3,4) |
|---|---|---|---|---|---|---|
| $\alpha_0$ | 0.929 | 0.858 | 0.951 | 0.855 | 0.948 | 0.806 |
| $\alpha_R$ | 0.929 | 0.858 | 0.951 | 0.855 | 0.948 | 0.806 |
| $\alpha_X$ | 0.944 | 0.833 | 0.944 | 0.858 | 0.948 | 0.790 |

**Table 6.** Accuracy of the selectivity analysis with three phases manipulated.

| $\alpha$ | (1,2,3) | (1,2,4) | (1,3,4) | (2,3,4) |
|---|---|---|---|---|
| $\alpha_0$ | 0.796 | 0.904 | 0.769 | 0.769 |
| $\alpha_R$ | 0.796 | 0.904 | 0.769 | 0.769 |
| $\alpha_X$ | 0.778 | 0.917 | 0.775 | 0.796 |

When three phases are modified, leaving only one phase not manipulated, the best performance was observed when only phase (3) remained original, achieving accuracy scores of 0.904 to 0.917 depending on the replacement value. While this is the highest accuracy among the tested configurations, it still represents a significant performance loss compared to the base model's accuracy of 0.959. The highest performance loss was again when the combination of phase (3) and (4) was manipulated with a score of 0.769.

Overall, the results of the selectivity analysis align with the expectations from the phase importance analysis, and therefore, it can be derived that the highlighted features indeed have a significant impact on the decision-making process of the fault detection.

*5.3 Qualitative evaluation of the visualisation*

In order to evaluate the visualisation, the most relevant phases for a specific quality class are identified by two process experts with a qualitative approach. The experts are presented with a crimp force curve, which is labelled with a specific quality class, and asked to identify the phases that most clearly indicate this quality class. They are also asked to describe the physical behaviour during the crimping process. For each quality class, the experts assigned a score ranging from 0 to 3 to each phase. Score 0 indicates that the segment had no impact or influence, score 1 a minor impact, score 2 a high impact, and score 3 a significant impact. The expert ratings are then compared with the visualisation output of the fault detection system. Table 7 and Table 8 show the expert ratings for the two quality classes Missing Strands and Crimped Insulation.

For the quality class of Missing Strands only minor discrepancies were observed in the ratings. While one expert considered phase (2) to have minor impact, the other expert rated the phase as not impactful. The experts concur that phase (1) and (4) is not influenced by this defect class. However, during phase (2), the force curve can exhibit a delayed increase compared to an OK connection, with the delay becoming more pronounced with an increasing number of missing strands. The effect can be influenced by various factors, such as the placement of the loose wires or the number of missing strands. In some cases, this effect may only become apparent in phase (3). Consequently, phase (2) is considered moderately influential by one expert, but not impactful by the other. Phase (3) is identified as the most impactful phase, as a reduced total force is applied during crimping when fewer strands are present.

In the context of Crimped Insulation discrepancies can be identified in the rating of phase (2) and (3). While one expert assigns a significant impact rating to phase (2) and (3), expert 2 designates them as having only high to minor importance. However, this can depend on various factors, including the wire cross-section of the conductor and the thickness of the insulation, as well as the degree of automation in the crimping process. For instance, greater insulation thickness may cause the force curve to increase earlier, whereas thinner insulation results in a less pronounced effect. During phase (2), the force curve deviates significantly from an OK curve, as the resistance introduced by the insulation adds to that of the conductor. Consequently, this area is considered highly impactful for detecting Crimped Insulation defects. In the transition to phase (3), the insulation material softens due to the heat generated by pressure. This may lead to a temporary decrease in the force curve, followed by a rise as the softened insulation material partially fills the voids within the individual strands. The evaluation of phase (3) varies significantly among experts, ranging from minor to very high influence, because the transition phase can be attributed to both phases, (2) and (3). Again, the experts concur that phase (1) and (4) is also not influenced by this defect class.

**Table 7.** Importance rating quality class Missing Strands.

| Process expert | (1) | (2) | (3) | (4) |
|---|---|---|---|---|
| #1 | 0 | 0 | 3 | 0 |
| #2 | 0 | 1 | 3 | 0 |

**Table 8.** Importance rating quality class Crimped Insulation.

| Process expert | (1) | (2) | (3) | (4) |
|---|---|---|---|---|
| #1 | 0 | 3 | 3 | 0 |
| #2 | 0 | 2 | 1 | 0 |

A comparison of these insights with the result of the visualisation, as depicted in Fig. 10, provides qualitative evidence that the visualisation of the fault detection systems highlights similar features for these two quality classes as the process experts. For the Missing Strands defect class (a), the visualisation assigns the highest level of importance to phase (3), with moderate support from phase (2) and a slight emphasis on phase (4). This overall distribution aligns well with the ratings provided by process experts. In contrast, the Crimped Insulation defect class (b) also highlights phase (3) as the most significant, with additional support from phase (4) and a slight influence from phase (1). While this general pattern aligns with expert assessments, some discrepancies emerge. The process expert noted that deviations from an OK curve can be observed in phases (2) and (3), depending on factors such as insulation thickness. Consequently, a shift in the rating within these two phases is plausible. Despite these variations, both the visualisation and expert evaluations agree that phases (1) and (4) play only a minor or negligible role. However, the data-driven model may identify additional interdependencies and, therefore, is more likely to slightly highlight these two phases. With regards to phases (2) and (3), both the process expert and the fault detection system conclude that these phases are the most relevant. Given that the effects on the force curve can be observed in both phases due to external factors, an overall agreement can be derived. Summarised, this demonstrates that the assessment of human experts confirms the model's findings, while the system contributes a more nuanced perspective by identifying subtle phase-level interdependencies that may not be immediately apparent to the human observer.

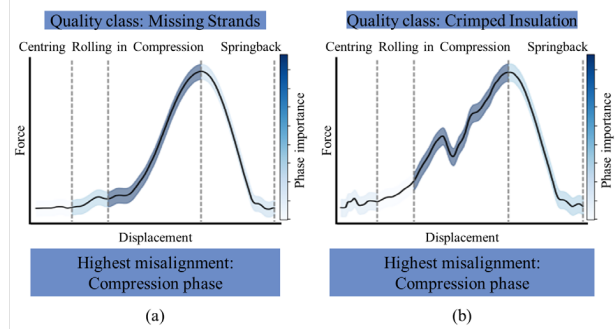

**Fig. 10.** Visualisation result of the fault detection system.

## 6. Conclusion

Modern manufacturing environments increasingly demand data-driven fault detection systems that are not only highly accurate but also interpretable, particularly in safety-critical applications. Conventional approaches offer transparency, but encounter challenges with the complexity and variability inherent in production data. Conversely, data-driven methods demonstrate superior fault detection performance but frequently lack interpretability, which hinders their practical adoption in industrial contexts. In order to overcome the identified trade-off, this study proposes a transparent, data-driven fault detection system that integrates three key components: firstly, a fault detection algorithm for multi-class classification of quality states; secondly, a model-agnostic explanation method for post hoc interpretability; and thirdly, a visualisation strategy designed for industrial usability. The methodology was validated in the domain of crimping using crimp force curves as univariate, discrete time series, but the approach can be adopted across other manufacturing domains. Furthermore, a combined quantitative and qualitative approach is proposed to evaluate the system. One fundamental aspect of the system is the utilisation of a subsequence-based explanation approach. Rather than attributing importance to isolated time series points, which are often uninterpretable in manufacturing contexts, the model aggregates SHAP values over physically meaningful curve phases. This mapping bridges the gap between statistical importance and domain expertise, enhancing the operator's ability to understand and act on model outputs. The resulting explanations are presented using a colour-coded overlay on the force curve, thus drawing the user's attention to the most influential segments. The proposed fault detection system achieved an overall accuracy of 95.9 %. The perturbation analysis demonstrated that manipulating the most relevant phases, identified by SHAP, resulted in the most significant decline in model performance, thereby confirming the validity of the explanatory insights. Additionally, expert assessments aligned closely with the visualised outputs of the system, further supporting its practical interpretability. Despite the evident benefits of the system the unequal distribution of data points across curve phases in this evaluation use case may influence the results of the selectivity analysis. In this context, future research should explore whether time series segmentation should be dynamically adapted for ML applications or remain aligned with established domain-specific conventions, as was done in this evaluation use case. Furthermore, the integration of process expertise through domain specific Large Language Models has the potential to further enhance human-centric data-driven quality control systems. The findings outlined, in conjunction with clear implications for both academic research and industrial practice, emphasise the relevance of this study.

**Acknowledgments**

The research presented in this paper is part of the project "Development of a machine learning based force curve analysis for a holistic process monitoring and quality assessment of crimp connections" (01IS22027D), which is funded within the initiative KI4KMU by the Federal Ministry of Education and Research (BMBF) and supervised by the project sponsor "German Aerospace Center (DLR) e.V.". Sincere appreciation is extended to the process experts at Schäfer Werkzeug- und Sondermaschinenbau GmbH, GKT Grüner Kabeltechnik and DD Kabelkonfektion Dropulic.

**Data availability**

The data used for this study is part of the openly available Crimp Force Curve Dataset at Harvard Dataverse: https://doi.org/10.7910/DVN/WBDKN6

**Declaration of Generative AI and AI-assisted technologies in the writing process**

During the preparation of this work the authors used ChatGPT in order to improve readability and language. After using this tool/service, the authors reviewed and edited the content as needed and takes full responsibility for the content of the publication.